\documentclass[letterpaper]{article}
\usepackage{notaaai}
\usepackage{times}
\usepackage{helvet}
\usepackage{courier}
\usepackage{tikz}
\usepackage{hyperref}
\usepackage{xcolor}
\usepackage{booktabs,chemformula}
\usepackage{tabulary}
\usepackage{tabularray}
\usepackage{array}

\hypersetup{
    colorlinks,
    linkcolor={red!50!black},
    citecolor={blue!50!black},
    urlcolor={blue!80!black}
}
\usepackage{siunitx}
\usetikzlibrary{shapes.geometric, arrows}
\tikzstyle{startstop} = [rectangle, rounded corners, 
minimum width=2.5cm, 
minimum height=1cm,
text centered, 
text width=2.5cm,
draw=black, 
fill=green!10]

\tikzstyle{io} = [trapezium, 
trapezium stretches=true, 
trapezium left angle=70, 
trapezium right angle=110, 
text width=2cm,
minimum width=2cm, 
minimum height=1cm, text centered, 
draw=black, fill=cyan!10]

\tikzstyle{process} = [rectangle, 
minimum width=3cm, 
minimum height=1cm, 
text centered, 
text width=2.5cm, 
draw=black, 
fill=yellow!10]
\tikzstyle{process2} = [rectangle, 
minimum width=3cm, 
minimum height=1cm, 
text centered, 
text width=3cm, 
draw=black, 
fill=blue!10]

\tikzstyle{startstop2} = [rectangle, rounded corners, 
minimum width=3cm, 
minimum height=1cm,
text centered, 
text width=5cm,
draw=black, 
fill=green!10]

\tikzstyle{startstop3} = [rectangle, rounded corners, 
minimum width=3cm, 
minimum height=1cm,
text centered, 
text width=5cm,
draw=black, 
fill=yellow!10]

\tikzstyle{startstop4} = [rectangle, rounded corners, 
minimum width=3cm, 
minimum height=1cm,
text centered, 
text width=5cm,
draw=black, 
fill=red!10]

\tikzstyle{decision} = [diamond, 
minimum width=2cm, 
minimum height=1cm, 
text width = 2cm,
text centered, 
draw=black, 
fill=orange!10]
\tikzstyle{arrow} = [thick,->,>=stealth]
\begin{document}
%
\title{AI for Anticipatory Action: Moving Beyond Climate Forecasting}
\author{Benjamin Q. Huynh\\
Department of Environmental Health \& Engineering\\ Johns Hopkins University\\
bhuynh@jhu.edu\\
\And
Mathew V. Kiang\\
Department of Epidemiology \& Population Health\\
Stanford University\\
mkiang@stanford.edu
}
\maketitle
\begin{abstract}
\begin{quote}
Disaster response agencies have been shifting from a paradigm of climate forecasting towards one of anticipatory action: assessing not just what the climate will be, but how it will impact specific populations, thereby enabling proactive response and resource allocation. Machine learning (ML) models are becoming exceptionally powerful at climate forecasting, but methodological gaps remain in terms of facilitating anticipatory action. Here we provide an overview of anticipatory action, review relevant applications of machine learning, identify common challenges, and highlight areas where ML can uniquely contribute to advancing disaster response for populations most vulnerable to climate change.\end{quote}
\end{abstract}
\section{Background}
Anticipatory action – proactive measures taken in response to forecasted risks and disasters before they occur – has emerged as a crucial framework for disaster response (Figure \ref{fig1}) \cite{weingartner2019anticipatory,thalheimer2022role}. Central to anticipatory action is impact-based forecasting, an approach that focuses on predicting the specific impacts of an impending disaster, such as health outcomes, disruption of services, or economic loss \cite{yu_impact-based_2022,merz2020impact,huynh2021public,liu2012progresses,gerl2016review}. By contrast, climate forecasting typically reports environmental conditions, such as wind speeds or rainfall. Impact-based forecasting is typically more accessible to humanitarian practitioners, but harder to predict; climate forecasting is generally more accurate, but may not adequately communicate potential hazards. However, the two approaches are not mutually exclusive – in fact, we examine here areas in which machine learning can uniquely be applied to leverage the predictive power of climate forecasting as well as the applied perspective of impact-based forecasting. 

Despite the promise of increased resilience from anticipatory action, major gaps remain in terms of developing robust, trustworthy systems. Forecast-based financing, a system where financing is allocated proactively based on risk estimates for models, has become increasingly used within United Nations agencies and non-governmental organizations. However, because regions that receive higher-risk forecasts receive more aid, poor predictive performance and causal validity of such forecasts may lead to misallocation of resources \cite{coughlan2015forecast}. The gap between the promise and current reality of anticipatory action may be in part aided by advances in AI research; we examine here potential areas of opportunity where AI can be used to improve anticipatory action.

\noindent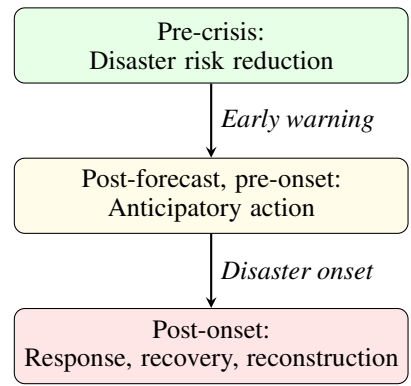
\begin{figure}[h]
\centering
\begin{tikzpicture}[node distance=2cm,
    >=stealth,
    node distance=2cm]
\node (in1) [startstop2] {Pre-crisis:\\ Disaster risk reduction};
\node (dec1) [startstop3, below of=in1] {Post-forecast, pre-onset: Anticipatory action};
\node (dec2) [startstop4, below of=dec1] {Post-onset: \\Response, recovery, reconstruction};

\draw [arrow] (in1) -- node[anchor=west] {\emph{Early warning}} (dec1);
\draw [arrow] (dec1) -- node[anchor=west] {\emph{Disaster onset}} (dec2);

\end{tikzpicture}
\caption{Typical timeline of disaster preparedness, anticipatory action, and response. \emph{Disaster risk reduction} refers to strengthening resilience before crises occur, and \emph{anticipatory action} refers to proactive response and resource allocation after an early warning system flags an incoming disaster.}
\label{fig1}
\end{figure}
\section{Challenges and Opportunities}
\begin{figure}[h]
\begin{tikzpicture}[node distance=2cm,
    >=stealth,
    node distance=3cm,
    database/.style={
      cylinder,
      cylinder uses custom fill,
      cylinder body fill=red!10,
      cylinder end fill=red!10,
      shape border rotate=90,
      aspect=0.25,
      draw
    }
  ]

\node (in1) [database] {Raw data};
\node (dec1) [decision, below of=in1, yshift=-0.5cm] {Modeling strategy?};

\node (pro2a) [process, below of=dec1, yshift=-0.5cm] {Impact forecast};

\node (pro2b) [process, right of=dec1, xshift=2cm] {Climate forecast};
\node (out1) [io, below of=pro2b] {Exposure-response function};
\node (out2) [startstop, below of=out1] {Population-specific\\impact forecast};

\draw [arrow] (in1) -- (dec1);
\draw [arrow] (dec1) -- node[anchor=east] {Correlational} (pro2a);
\draw [arrow] (dec1) -- node[anchor=south] {Causal} (pro2b);
\draw [arrow] (pro2b) -- node[anchor=east] {$P = g(C|C = c)$} (out1);
\draw [arrow] (out1) -- node[anchor=east]{$Y = \phi(P,V|P=p,V=v)$} (out2);
\draw [arrow] (pro2a) |- node[anchor=north]{$Y = f(C,V|C = c, V = v)$}(out2);
\end{tikzpicture}
\caption{Typical workflow for the forecasting portion of anticipatory action. After population-specific impact forecasts are provided, targeted resource allocation can be conducted.}
\label{workflow}
\end{figure}
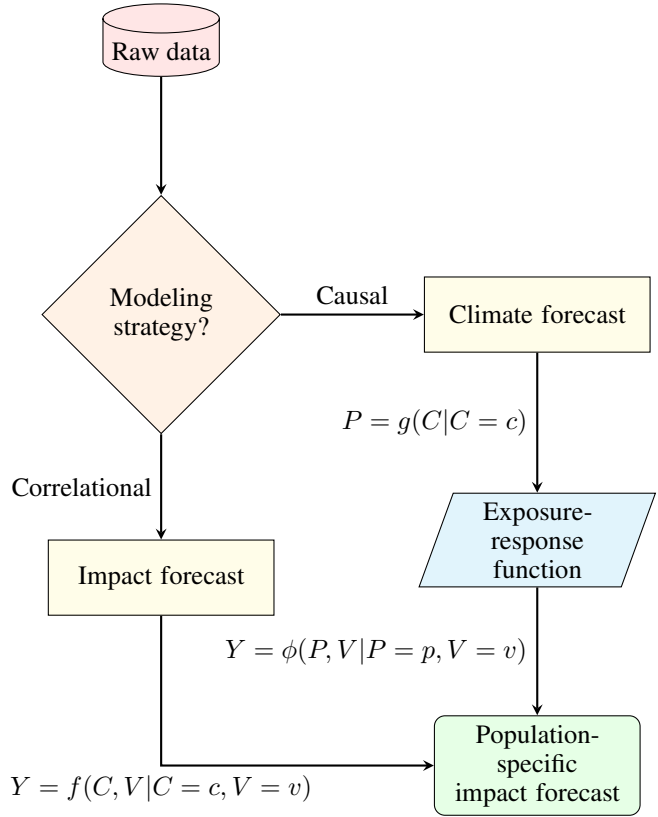

Figure \ref{workflow} shows a typical undertaking of the forecasting portion of anticipatory action. A model is developed based on spatiotemporal environmental data and used to make population-specific impact forecasts. From there, the forecasts are used to enable proactive resource allocation. We broadly identify two strategies to forecast disaster impacts: (1) a correlational forecasting approach where the impacts of the disaster are directly modeled; and (2) a causal approach wherein the disaster is meteorologically modeled, then outputs are run through an exposure-response function to ascertain the impact of the disaster. A subsequent step is taken after either approach to apply the model to a target population for a population-specific forecast. Table \ref{tab1} shows examples of how impact-based forecasts can be used.
\noindent
\subsection{Problem Formulation}
Formally, let $C = c_{1}, c_{2}, ... c_{k}$ be a set of climate-related features and let $V = v_{1}, v_{2}, ... v_{k}$ be a set of population-specific features. The correlational forecasting approach to directly model disaster impacts is represented by \[Y_{s,t+n} = f(C,V|C = c_{s,t}, V= v_{s,t})\] where $Y$ is the outcome of interest, $s$ and $t$ are spatiotemporal indices, and $n$ is the forecasting lead time. 
Machine learning approaches have been used for impact-based forecasting for outcomes related to disasters, such as modeling displacement, food insecurity, disease outbreaks, economic loss, or infrastructure assessment \cite{pham_predictive_2022,huynh_forecasting_2019,foini2023forecastability,wang2019defsi,proma_nadbenchmarks_nodate,oshri2018infrastructure}. Most ML approaches for impact-based forecasting directly model the impact without using climate forecasting as an intermediary; to the extent climate variables are used, they are typically just more predictors for the model.

In terms of climate forecasting, such models can be represented as \[P_{s,t+n} = g(C|C = c_{s,t})\] where $P$ is a modeled climate event, such as a risk score or climate condition, and $g$ is the climate forecasting function. Machine learning models for climate forecasting, while historically inferior to global simulation methods like numerical weather prediction, have since reached comparable or even superior performance to simulation-based methods for a variety of forecasting applications, ranging from tropical cyclones, heat waves, rainfall, floods, wildfires, and more \cite{chen2020machine,mosavi2018flood,cifuentes2020air,jain2020review,zhang_skilful_2023,bi_accurate_2023,rolnick2022tackling}. ML models are far less computationally expensive than global simulation models, and can be used to supplement them by emulating their predictions. Because such models are emulating mechanistic physical processes, predictive performance is often very high compared to other forecasting domains. Many ML models frame their outputs as risk scores, providing spatiotemporal risk estimates for a given type of natural disaster. 

Causally forecasting disaster impacts is represented by \[Y_{s,t+n} = f(C,V |do(C = c_{s,t}),V=v_{s,t})\] where the $do$ operator refers to causally setting the climate variables $C$ to be specific values, rather than using a correlational relationship \cite{pearl2009causality}. In practice, causal forecasting is facilitated by an \emph{exposure-response function}: \[Y = \phi(P,V|P=p,V=v)\] where $\phi$ represents the relationship between the climate event and the impact of interest. Exposure-response functions are typically used in environmental epidemiology, where the relationship between an exposure and a health outcome is identified, then that relationship is used to simulate health outcomes based on known exposure levels \cite{aunan2004exposure}. Depending on how they are identified, exposure-response functions can vary in levels of causal validity, ranging from results from randomized controlled trials to associations identified from observational data using causal inference. In other fields, this relationship may also be known as a \emph{damage function} \cite{merz2020impact}.
\begin{table*}
\caption{Example scenarios of disasters and their potential forecasts.}
\begin{tabulary}{\linewidth}{m{1cm}m{4cm}m{5.5cm}m{5.5cm}}
\toprule
\textbf{Hazard}           & \textbf{Climate Forecast}                                                                                     & \textbf{Exposure-response function(s)}                                                                                                & \textbf{Impact-based Forecast}                                                                                                                                 \\ \midrule
Tropical cyclone & A category 4 tropical cyclone with windspeed 220km/h is anticipated to make landfall in 36-48 hours. & Relationship between tropical cyclone and building damage/displacement                                                       & 40-60\% of housing in Region A will be damaged, with an anticipated 20,000-40,000 displaced individuals.                                        \\ \midrule
Wildfire smoke   & Wildfire smoke will reach City B tomorrow morning and cover the city for 48-72 hours.                & Relationship between smoke exposure and hospitalizations; relationship between excess hospitalizations and drug shortages    & 200-400 excess hospitalizations are expected, and extra inhalers will be necessary to prevent shortages.                                          \\ \midrule
Drought          & Precipitation far below average is likely for the summer in Province C.                                            & Relationship between precipitation and crop failure/food insecurity, relationship between food insecurity and malnourishment & Food insecurity will increase, and prices for affected crops will spike by 55-75\%. Incidence of child malnourishment will increase by 20-30\%. \\ \bottomrule
\end{tabulary}
\label{tab1}
\end{table*}
\subsection{Open Areas for Improvement}
The decision to use a causal or correlational modeling strategy is context-specific, and both approaches have their own advantages and disadvantages. For causal forecasting, exposure-response functions can be difficult to identify, depending on the exposure.
Correlational forecasting is subject to confounders, threatening causal validity \cite{peters2016causal}. Both approaches face challenges of dataset shift and algorithmic fairness. Forecasting for anticipatory action thus remains an open problem with multiple potential areas for improvement, and we discuss here how AI can be used to address such issues.
\subsubsection{Exposure-response functions}
Exposure-response functions are difficult to obtain, and especially so for disaster-related impacts. Disasters vary in magnitude, location, and population characteristics, with relatively infrequent occurrence -- estimating a valid association between a disaster and an impact typically requires high-resolution data and a plausible empirical strategy \cite{parks2022association}. Even air pollution, which is continuously monitored with high-resolution exposure and health outcomes data, drives contention over what the exposure-response function should be \cite{zigler2014point}. Machine learning can be useful in improving estimation of exposure-response relationships by producing better characterizations of disaster impacts. For example, satellite imagery in conjunction with AI has been used to assess building damage from disasters \cite{novikov2018satellite,zhao2020building,oshri2018infrastructure}; such information could be used to develop exposure-response relationships between disaster occurrence and building damage, potentially enabling forecasts of building damage.

As another example, AI can be used to improve estimates of health outcomes such as mortality or disease prevalence after a disaster. In low-resource environments, post-disaster estimates of mortality or disease prevalence are conducted by taking surveys across clusters of households. Results from such surveys can vary drastically due to sampling bias, which can be difficult to overcome through randomization due to poor access to households post-disaster. The survey approach remains the gold-standard, as official national or international statistics are generally too slow to capture the magnitude of a disaster shortly after it has occurred \cite{kishore2018mortality,gang2023cross}. AI can be used in conjunction with post-stratification, a technique typically used in the social sciences to learn from biased samples, to estimate adjusted mortality or disease prevalence rates \cite{gelman2016mythical,broniecki2020improved,van2014association}.

Additionally, mortality or disease prevalence surveys typically obtain population-wide estimates by extrapolating from survey results: \[
\frac{E_s}{N_s} \approx \frac{E_f}{N_f}
\] where $E$ and $N$ represent events and population size respectively, and $s$ and $f$ indices indicate observations from the survey data and the full population at large, respectively. Typically, $E_{f}$ is estimated by assuming a value for $N_{f}$ (as well as adjusting for any covariates), but there is often uncertainty regarding the value of $N_{f}$ as populations tend to migrate in response to disasters, potentially affecting estimates of $E_{f}$. Any AI methods to estimate $N_{f}$, whether through migration or population modeling, can thus be valuable in improving identification of exposure-response functions \cite{doocy2013mortality,robinson2017deep,yeh2020using}.

\subsubsection{Dataset Shift}
All forecasting approaches are subject to dataset shift, where there is a difference in distribution between training data and deployment-time testing data. This is particularly true for disaster-related forecasts: first, unprecedented events happen regularly under climate change, and models may not provide valid results for environmental conditions that have never been seen before \cite{dickerman2020counterfactual}; secondly, because disasters happen relatively infrequently, there will not be sufficient historical data regarding how a disaster of any given magnitude affects any given region or subpopulation of a country. Taken together, high-resolution country-wide forecasts will require targeted population-specific forecasts in a manner that addresses dataset shift.

One approach to handling dataset shift is leveraging \emph{transportability} of causal effects \cite{bareinboim2012transportability,pearl2022external}. Transportability provides a framework to assess whether causal effects can be applied from one environment to another. As an example, suppose we have a plausible causal effect estimate saying that wildfire smoke caused an $X\%$ increase in hospitalizations per \si{\micro\gram}/\si{\cubic\metre} of fine particulate matter in City A. We want to know how many additional hospitalizations we'd expect to see if smoke reached City B, but City B has a different age distribution than City A, and age is a confounder in that it affects both exposure levels to smoke and likelihood of hospitalization. Transportability theory determines that if we can identify both the full causal effect: \[P(hospitalization|do(smoke))\] and the age-specific causal effect: \[P(hospitalization|do(smoke),age)\] then it is feasible to transport causal estimates from City A to City B. Transportability has been applied in various complex scenarios in health \cite{rudolph2018composition,prosperi2020causal}, and could also be useful for improving causal validity of disaster forecasts.

Relatedly, another approach to handle dataset shift for impact forecasting involves \emph{invariant prediction} \cite{peters2016causal,pfister2019invariant,arjovsky2019invariant}. Briefly, approaches leveraging invariant prediction allow for selection of causal features from a model by identifying model configurations with high prediction invariance, enabling more robust predictions less likely to be influenced by spurious correlations. Using approaches based on invariant prediction would improve the causal validity of correlational impact forecasting without requiring the intermediary step of calculating an exposure-response function, thereby improving the reliability of impact predictions in relationship to varying climate variables. Although these approaches require the strong assumption of there being no unobserved confounders present, sensitivity analyses for potentially unobserved confounders could be used to provide uncertainty intervals for predictions \cite{rosenbaum1983assessing,yadlowsky2022bounds,jung2017simple}.
Applications of invariant causal prediction have recently been gaining traction within medicine \cite{prosperi2020causal,subbaswamy2019preventing,subbaswamy2022unifying} and earth sciences \cite{runge2019inferring,sheth2022stcd}, suggesting that applications to climate disasters and anticipatory action may be feasible. 

\subsubsection{Data Scarcity and Algorithmic Bias}

Climate data are inequitably collected, with marginalized populations typically having poorer and lower-quality data \cite{rolnick2022tackling}. 60\% of countries lack basic water information services that can be used for flood preparedness, and roughly half of all countries do not have multi-hazard early warning systems \cite{waterreport}. The World Meteorological Organization roughly estimates that countries without such early warning systems have over eight times the disaster-related mortality rates than countries with early warning systems \cite{honore2022global}.
Data scarcity degrades the performance of predictive models for regions that are most vulnerable to disasters in the first place, making anticipatory action more difficult within low-resource countries. Typical numerical weather prediction models rely on data collected globally to make better localized predictions, but AI efforts to ameliorate data scarcity such as localized predictions via transfer learning, or up/downscaling climate predictions may help circumvent such requirements and improve impact-based forecasts \cite{xie2016transfer,li2021improved,vosper2023deep}.

Careful consideration needs to be given to the ethical implications of AI models for impact forecasting, particularly as forecast-based financing becomes increasingly used \cite{coughlan2015forecast}. In particular, practitioners should exercise caution as to how the choice of metric affects equity considerations. Treating economic loss as an absolute metric for impact assessment favors wealthier regions; considering risk and loss as relative has been shown to provide more equitable results in risk assessment \cite{kind2020social}. Additionally, considering how disasters impact regions beyond economic loss, and instead using metrics like health or wellness, may provide a more nuanced characterization of disaster impacts \cite{hino2021five}. Algorithmic audits and assessments commonly used for composite indicators or risk scoring machine learning models should be used to assess the algorithmic fairness of impact-based forecasts \cite{mayson2019bias,obermeyer2021algorithmic,huynh2023potential,bhagwat2023mitigating,paulus2020predictably}.

\section{Discussion}
In this paper, we surveyed how AI advances can be used to improve anticipatory action in terms of current methodological gaps and potential areas for improvement. There are several limitations to our paper. First, this is not an exhaustive review of all relevant literature: we refer to other review papers where applicable, and subfields such as AI for climate forecasting or causal prediction are sufficiently deep and nuanced that reviewing all relevant works is out of scope. Secondly, there are other crucial areas of improvement that are not mentioned here, such as model explainability or uncertainty quantification -- our subjective judgment is that they represent long-standing challenges of fundamental importance that are not sufficiently specific to anticipatory action to be in scope of this survey.

Amidst intensifying climate change and increased frequency of natural disasters, the need for improved anticipatory action will only rise. Some of the AI-related challenges and opportunities we raise may appear too intractable or open-ended. We argue that since anticipatory action is already being widely used and deployed, any improvements to existing systems, however marginal, could be consequential and life-saving. Such a perspective could be used to foster a long-term vision for a future where we have better data and tools, enabling protection of the most vulnerable from disasters before they occur.

\bibliographystyle{aaai} 
\bibliography{bib}

\end{document}